# Using the profile of publishers to predict barriers across news articles

Abdul Sittar[1,2][0000−0003−0280−9594] and Dunja Mladenić[1,2][0000−0002−0360−6505]

[1] Jožef Stefan Institute, Slovenia,
[2] Jožef Stefan International Postgraduate School, Slovenia,
Jamova cesta 39
{abdul.sittar, dunja.mladenic}@ijs.si

**Abstract.** Detection of news propagation barriers, being economical, cultural, political, time zonal, or geographical, is still an open research issue. We present an approach to barrier detection in news spreading by utilizing Wikipedia-concepts and metadata associated with each barrier. Solving this problem can not only convey the information about the coverage of an event but it can also show whether an event has been able to cross a specific barrier or not. Experimental results on IPoNews dataset (dataset for information spreading over the news) reveals that simple classification models are able to detect barriers with high accuracy. We believe that our approach can serve to provide useful insights which pave the way for the future development of a system for predicting information spreading barriers over the news.

## 1 Introduction

The phenomenon of event-centric news spreading due to globalization has been exposed internationally [8]. International events capture attention from all corners of the world. News agencies play their part to bring our attentions on some events and not on others. Varying nature of living styles, cultures, economic conditions, time zone, and geographical juxtaposition of countries present a significant role in process of publishing news related to different events [3, 6, 13, 19–21]. For example, publishing about sports events could be dependent on culture, epidemic events can reach firstly to neighboring countries due to geographic proximity and, news on a luxury product may be relevant for economically strong countries due to demand of wealthy people. We represent this differentiation along with different barriers. These barriers include but are not limited to 1) Economic Barrier, 2) Cultural Barrier, 3) Political Barrier, 4) Geographical Barrier, and 5) Time Zone Barrier. Detection of the overpass of these barriers does not only tell us the area where the broadcasting of an event reached, but it also shows us events-location relation as countries have different culture, economic conditions, geographical placement on the globe, political point of view, and time zone. Following are the definitions of news crossing these barriers:

**Cultural Barrier.** If we identify the coverage of specific event-centric news by publishers that are surrounded by different cultures, then we can say that the news related to the event crossed cultural barriers.

**Political Barrier.** If news about a specific event is disseminated from publishers having different political alignment, we can say that the news related to that event crossed the political barrier.

**Geographical Barrier.** We say that some news related to a specific event overpasses geographical barriers if that event gets attention by publishers of countries located in different geographical regions.

**Time Zone Barrier.** We can claim that event-centric news has crossed the time zone barrier if it has been published by publishers located in different time zones.

**Economic Barrier.** It can be asserted that a piece of event-centric news has crossed economic barriers if it is published in countries having different economic conditions.

In this paper, we propose a methodology for detection of different barriers during information propagation in form of news that utilize data (IPoNews) [18] related to three contrasting events (earthquake, Global warming, and FIFA world cup) in different domains (natural disasters, climate changes, and sports) in 5 different languages: English, Slovene, Portuguese, German, and Spanish.

### 1.1 Contributions

Following are the main scientific contributions of this paper:

- A novel methodology for barrier detection in news spreading.
- Experimental comparison of several simple classification models that can serve as a baseline.

### 1.2 Problem Statement

Observing the spreading of news on a particular event over time, we want to predict whether a barrier (cultural, political, geographical, time zone, economical) is likely to hamper information while information propagates over the news (binary classification).

## 2 RELATED WORK

Multiple barriers come across event-centric news specifically when the news is concerned about international or national events. According to news flow theories, multiple determinants impact international news spreading. The economic power of a country is one of the factors that influence news spreading. Moreover, economic variations has different influence for different events (e.g. protests, conflicts, disasters) [15]. The magnitude of economic interactivity between countries can also impact the news flow [21]. Economic growth/income level shows the economic condition of a country. Multiple organizations are working on generating prosperity and welfare index on yearly basis. Among them, “The Legatum Prosperity Index” and “Human Development Index” are popular [1], [2]. Geographical representation of entities and events has been utilized extensively in the past to detect local, global, and critical events [3, 13, 19, 20]. It has been said that countries with close distance share culture and language up to a certain extent which can further unfold interesting facts about shared tendencies in information spreading [15, 16].

News agencies tend to follow the national context in which journalists operate. One of the related examples is the SARS epidemic study which found that cross-national contextual values such as political and economic situations impact the news selection [5]. It will be true to say that fake news is produced based on many factors and it is surrounded by a paramount factor that is political effect [11]. A great amount of work regarding fake news dwells on different strategies and few studies considered political alignment to have a compelling effect on news spreading [4, 12]. [12] strongly proved it to be a major strategy in news agencies to control the news and change accordingly due to the involvement of journalists and political actors. Countries that share common culture are expected to have heavier news flow about between them reporting on similar events [21]. Many quantitative studies found demographic, psychological, sociocultural, source, system, and content-related aspects [1]. Many models have tried to explain cultural differences between societies. Hofstede’s national culture dimensions (HNCD) has been widely used and cited in different disciplines [7, 9].

News classification for different kinds of problems is a well-known topic since the past and features used to classify varies depending upon the problem. [17] used news content and user profile to classify the news whether it is fake or not. [2] calculated TF-IDF score and Word2Vec score of most frequent words and used them as features to classify into one of the five categories (state, economy, entertainment, international, and sports). Similarly, [14] performed part-of-speech (POS) tagging at sentences level and used them as features, and built supervised learning classifiers to classify news articles based on their location. Mostly classifier trained to utilize popular supervised learning methods such as Random Forest, Support Vector Machine (SVM), Naive Bayes, k-Nearest Neighbour (kNN), and Decision Tree. In this work, we used the profile of each barrier for each news publisher (see section 3.5) and most frequent 300 Wikipedia concepts from the dataset that appeared in the list of news articles related to three contrasting events (earthquake, Global Warming, and FIFA world cup). We also

1 http://hdr.undp.org/en/content/human-development-index-hdi
2 https://www.prosperity.com/

compared the results of popular classifiers such as SVM, Random Forest, Decision Tree, Naive Bayes, and kNN (see Section 5.4).

# 3 DATA DESCRIPTION

## 3.1 Dataset

We utilized dataset ”A dataset for information spreading over the news (IPoNews)” that consists of pairs of news articles that were labeled based on the level of their similarity, as described in [18]. This dataset was collected from Event Registry, a platform that identifies events by collecting related articles written in different languages from tens of thousands of news sources [10]. The similarity score among cross-lingual news articles was calculated using concept-based similarity employing Wikifier service[3]. [18] describes the criteria when information is considered to be propagated. Statistics of the data set are shown in table 3.

**Table 1.** Statistics about dataset

| Dataset | Domain | Event type | Articles per Language | | | | | Total Articles |
|---|---|---|---|---|---|---|---|---|
| | | | Eng | Spa | Ger | Slv | Por | |
| 1 | Sports | FIFA World Cup | 983 | 762 | 711 | 10 | 216 | 2682 |
| 2 | Natural Disaster | Earthquake | 941 | 999 | 937 | 19 | 251 | 3147 |
| 3 | Climate Changes | Global Warming | 996 | 298 | 545 | 8 | 97 | 1944 |

The dataset contains a list of pairs of news articles annotated with one of the labels such as ”information-Propagated”, ”Unsure”, or ”Information-Not-Propagated” (see Table 2). The information is considered to be propagated if the cosine similarity score of the two articles in the pair is above a predefined threshold ( $\geq 0.7$ for Information-Propagated, $< 0.4$ for Information-not-Propagated, otherwise Unsure). We restructured the original dataset to include only examples labeled as spreading information. In this way, we have pair of news articles where we observe information spreading from one to the other. Furthermore, for each example, instead of having a pair of articles, we kept only the article that was published earlier. In this way, each example contains an article that spreads information.

**Table 2.** Articles with metadata

| from | to | weight | Class | from-publisher | to-publisher | from-pub-uri | to-pub-uri |
|---|---|---|---|---|---|---|---|
| Por44 | Por43 | 0.627 | Unsure | ClicRBS | SAPO 24 | jornald.clicrbs.com.br | 24.sapo.pt |
| English881 | English880 | 1 | Information-Propagated | Sky News | 247 Wall St. | news.sky.com | 247wallst.com |
| English258 | English329 | 0.313 | Information-Not-Propagated | Sify | 4-traders | sify.com | 4-traders.com |
| English793 | English787 | 0.238 | Information-Not-Propagated | Bioengineer.org | 7NEWS Sydney | scienmag.com | 7news.com.au |
| German237 | German236 | 0.979 | Information-Propagated | watson | watson | aargauerzeitung.ch | aargauerzeitung.ch |

[3] http://wikifier.org/info.html, https://github.com/abdulsittar/IPoNews

### 3.2 Statistics after restructuring the data

The original dataset describes in Section 3 contains pairs of articles along with the information on whether there was the propagation of information related to a specific event or not. We used only examples labeled as propagating information [4]. Based on the available metadata for articles, we ignored articles that do not have metadata information in our database (see Section 3.4). Table 3 shows the statistics for each barrier after filtering the original dataset.

**Table 3.** Statistics about barrier

| Dataset | Domain | Event type | Articles for each barrier | | | | |
|---|---|---|---|---|---|---|---|
| | | | Time-Zone | Cultural | Political | Geographical | Economical |
| 1 | Sports | FIFA World Cup | 724 | 699 | 143 | 726 | 634 |
| 2 | Natural Disaster | Earthquake | 1102 | 1113 | 227 | 1113 | 1010 |
| 3 | Climate Changes | Global Warming | 586 | 445 | 108 | 487 | 463 |

### 3.3 Wikipedia Concepts as Features

As our dataset already mention (see Section 3) if information in news is spreading from an article to another based on Wikipedia-concepts, we utilized the most frequent (top 300) Wikipedia-concepts as features. Figure 1 portrays these Wikipedia-concepts for all three events in form of word clouds.

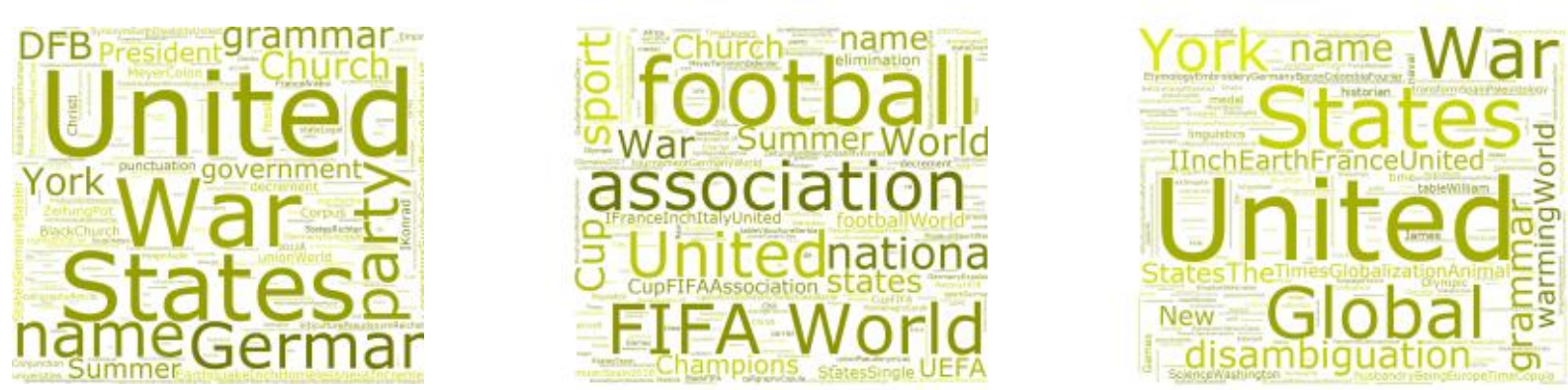

**Fig. 1. Word clouds of most frequent words related to earthquake, FIFA World Cup and Global Warming events respectively.**

### 3.4 Barriers Knowledge

Barriers knowledge refers to a database that contains metadata about each barrier. Figure 3 shows schema of database and Table 4 presents barriers along with their characteristics. Each barrier depends on one main information that is the country name of the headquarter of the news publishers. Since the utilized data

4 https://doi.org/10.5281/zenodo.3950064

set already contains headquarter of publishers therefore we fetched the country associated with headquarters. For economical barrier, we fetched economical profile for each country using “”The Legatum Prosperity Index”” [5]. Cultural differences among different regions were collected using Hofstede’s national culture dimensions (HNCD). For time zone and geographical barrier, we stored general UTC-offset, latitude, and longitude. For political barrier we are using the political alignment of the newspaper/magazine that we determined based on Wikipedia infobox at their Wikipedia page. For instance, for Austrian newspaper ”Der Standard” we find *social liberalism* as political alignment (See Figure 2), for British newspaper ”Daily Mail” we find *right-wing* as political alignment, for German ”Stern” magazine there is no information in its Wikipedia infobox on the political alignment thus we label political alignment as *unknown*.

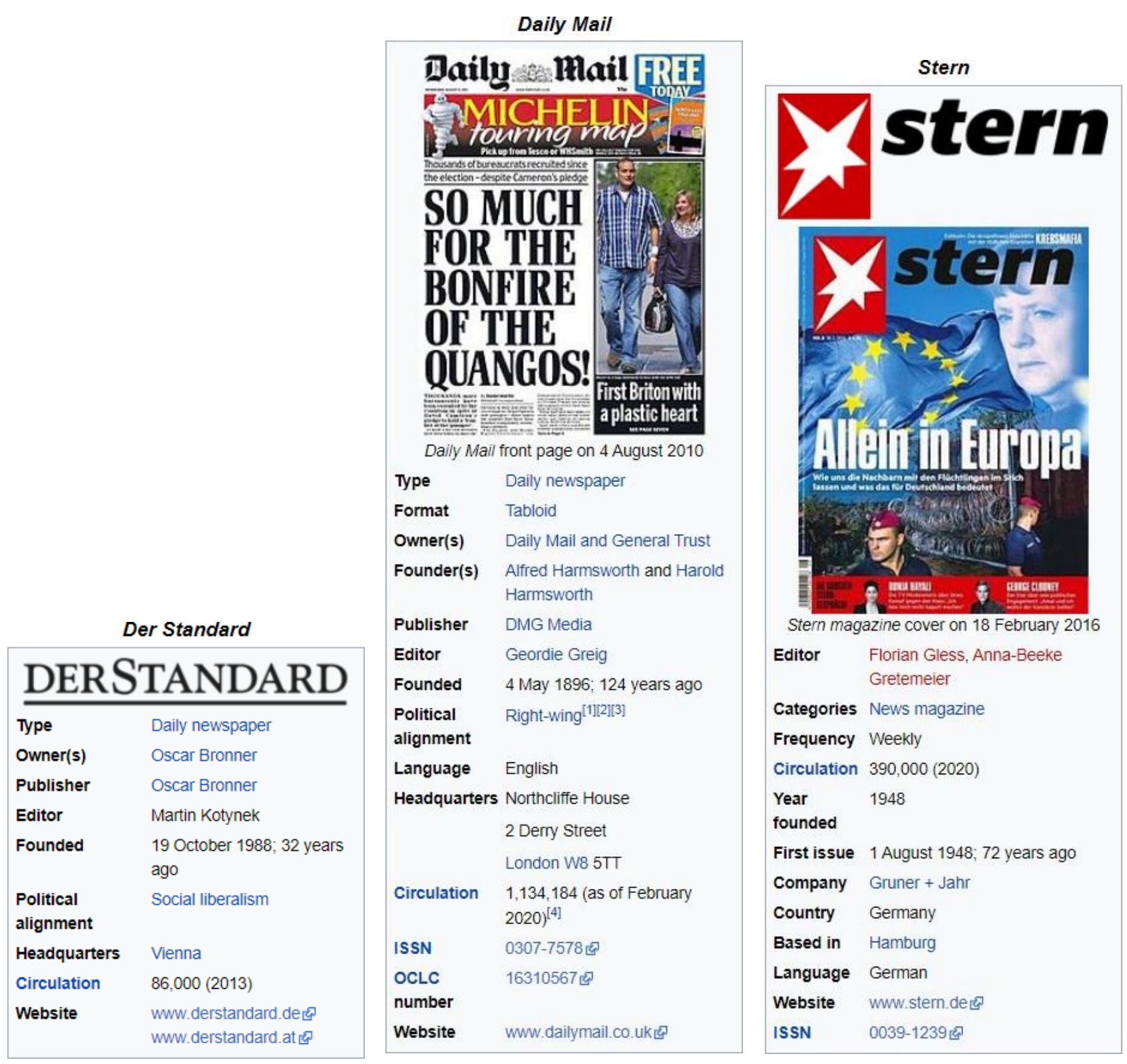

**Fig. 2. Three Wikipedia infobox for three different newspapers/magazines with political alignment**

5 https://www.prosperity.com/

### 3.5 Features for Individual Barrier

We represented each barrier with a specific profile containing a list of features. Table 4 depicts the list of features for each barrier. Economic and cultural barriers consist of a vector of length 11 and 6 features whereas geographical, time zone, and political only contain 1 or 2 features such as latitude-longitude, UTC-offset, and political alignment.

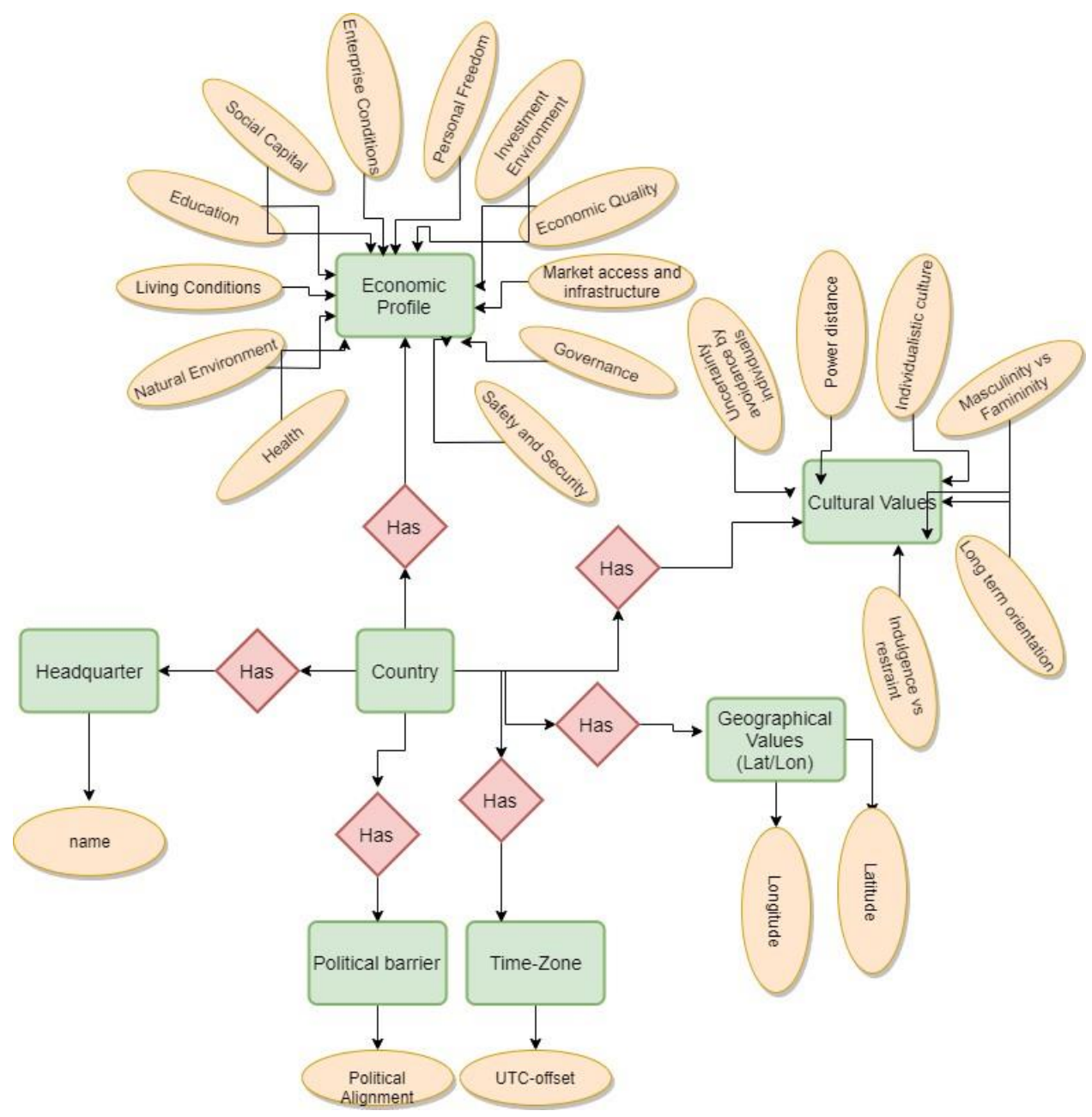

**Fig. 3.** Database Schema for Barriers

### 3.6 Dataset Annotation

We queried the metadata information for each article and generated a CSV file for each barrier. We annotated each article based on that meta information to be used for model training and classification. For economic and cultural barriers, we calculated cosine similarity between vectors of economical values and vectors of cultural values. Score greater than the threshold value of 0.9 labeled as FALSE otherwise TRUE. We set the lowest value as a threshold based on the fact that if two countries have a little gap concerning culture or economical values then there exists a barrier. For geographical barriers, we compared the latitude and longitude of the country of each publisher. If a country name or lat/lat appeared to be the same then we annotated it with FALSE otherwise TRUE. Lastly, for time-zone and political barriers, we followed the same process that was for the geographical barrier. if political alignment or UTC-offset appeared to be the same for a pair then it is annotated with FALSE otherwise TRUE. Figure 4 depicts the class distribution for each barrier. We can notice unbalanced class distribution with majority of the examples being False. This is especially true for Cultural and Political barrier with 91 percent of example being False. Thus in our evaluation we rely more on F1 measure than classification accuracy.

**Table 4.** Features of each barrier

| Barrier | Features |
|---|---|
| **Economic** | Rank, Safety-Security, Personal-Freedom, Governance, Social-Capital, Investment-Environment, Enterprise-Conditions, Market-Infrastructure, Economic-Quality, Living-Conditions, Health, Education, Natural-Environment |
| **Cultural** | Power-Distance, Uncertainty-Avoidance-By-Individuals, Individualistic-Cultures, Masculinity-Femininity, Long-Term-Orientation, Indulgence-Restraint |
| **Geographical** | Latitude, Longitude |
| **Time Zone** | UTC-offset |
| **Political** | Political-Alignment |

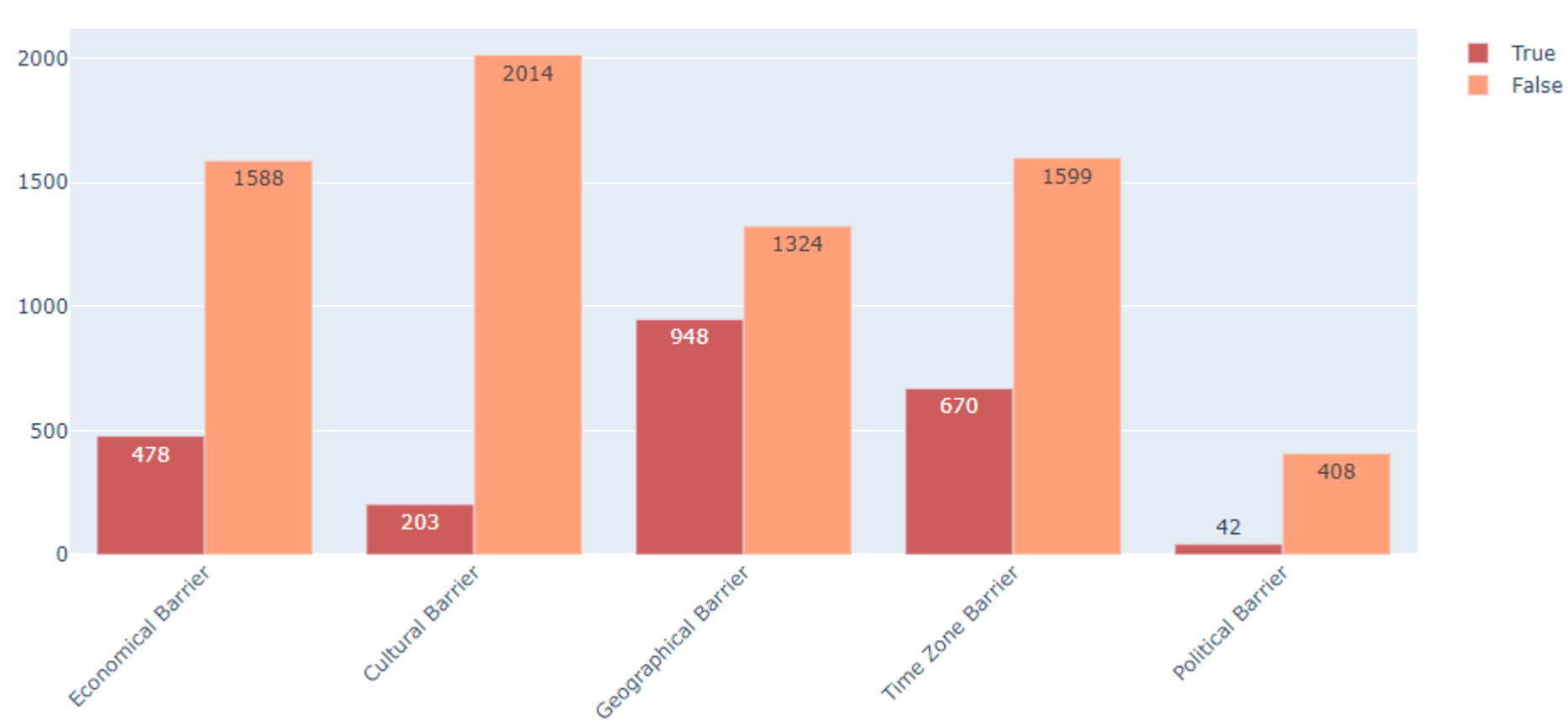

**Fig. 4.** **Class Distribution for Each Barrier**

# 4 MATERIALS AND METHODS

## 4.1 Problem Modeling

For each barrier, we have a list of news articles where each article is associated with 300 Wikipedia-concepts and features related to that barrier. The task is to predict the status S of each barrier B.

$$S = f(C, B)$$

f is the learning function for barrier detection, C is donating here Wikipedia-concepts related to an article and B is the list of features related to a specific barrier (see Table 4).

## 4.2 Methodology

We utilized dataset IPoNews [18] and built a database on top of this dataset that includes barrier knowledge. Figure 5 explains the overall process of model construction from news articles to results generation. We created a list of instances using the most frequent Wikipedia-concepts based on news articles and joined them along with barrier knowledge. After performing the annotation (see Section 3.6), we trained popular classification models and generated the results on test data (see Section 5.4).

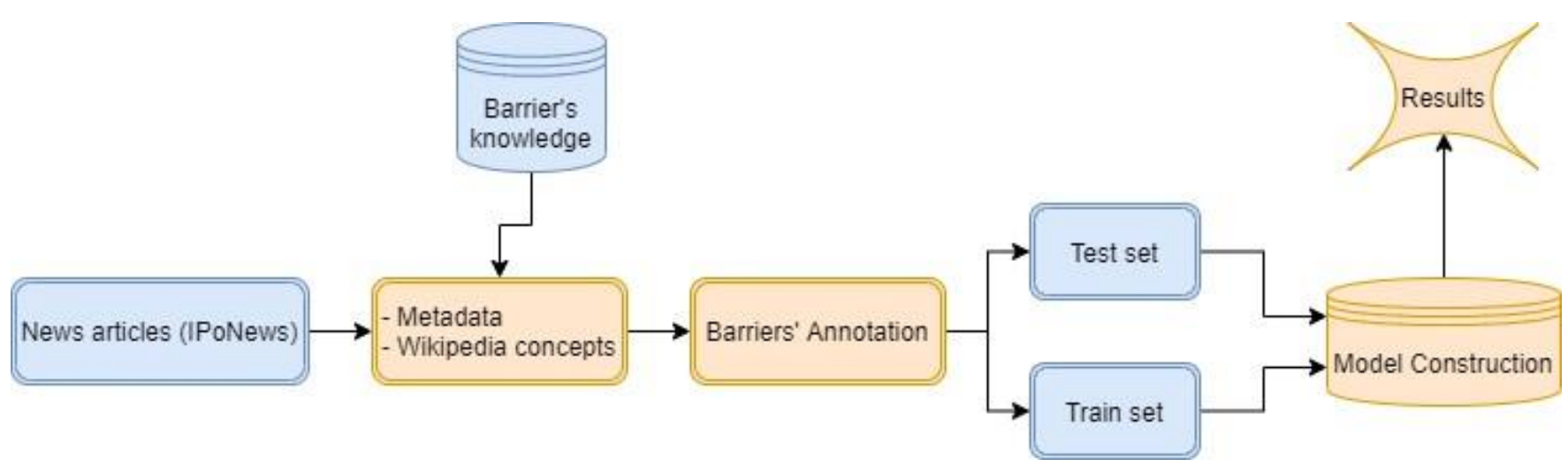

**Fig. 5.** Steps for Model Construction

# 5 EXPERIMENTAL EVALUATION

## 5.1 Baselines

We used the following methods as baselines for all our models.

- **Uniform:** Generates predictions uniformly at random.
- **Stratified:** Generates predictions by respecting the training set's class distribution.
- **Most Frequent:** Always predicts the most frequent label in the training set.

### 5.2 Classification Methods

We trained popular classification models for each barrier such as SVM, kNN, Decision Tree, Random Forest, and Naive Bayes using Scikit-Learn. We applied a stratified 10-fold cross-validator to split the dataset for training and testing. For Random Forest, kNN, and Decision Tree, we varied the size of n-estimator, value of k, and max-leafs and chosen the one with the best score on test data respectively. Implementation of this methodology to barrier detection can be found on GitHub [6].

### 5.3 Evaluation Metric

Due to imbalance in the class distribution for all barriers, we used micro averaged precision and recall to evaluate our models. [7]

- **Micro-Precision:** The precision of average contributions from each class is calculated in micro-precision whereas the following question is answered by precision: What proportion of positive predictions was correct? It is defined as:

$$Micro - Precision = \frac{TruePositive_{sum}}{TruePositive_{sum} + FalsePositive_{sum}}$$

- **Micro-Recall:** Recall of average contributions from each class is calculated in micro-recall whereas the following question is answered by recall: What proportion of actual positives was predicted correctly? It is defined as:

$$Micro - Recall = \frac{TruePositive_{sum}}{TruePositive_{sum} + FalseNegative_{sum}}$$

### 5.4 Results and Analysis

Table 5 shows the results of all the classifiers for each barrier along with baselines. Analysis of the experimental results show that overall all the machine learning models outperform the three baselines. For all the barriers, we can notice Micro-Recall is equal to Micro-Precision. The best performing baseline is the ”Most-frequent” with Micro-F1 for economic, cultural, geographical, time zone, and political barrier equal to 0.70, 0.90, 0.58, 0.70, and 0.90 respectively. The best performing models on all the barriers are Decision Tree, Random Forest, and kNN. Looking at Micro-F1, we can see that on the Economic and Cultural barrier kNN achieved the best performance of 0.75 and 0.95 respectively. On Geographical barriers, kNN and Decision Tree performed the best achieving 0.81. On Time-Zone, the best performing classifier is Random Forest with Micro-F1

[6] https://github.com/cleopatra-itn/BarrierDetection-Classification

[7] https://peltarion.com/knowledge-center/documentation/evaluation-view/classification-loss-metrics/micro-recall

0.83. On Political barriers, SVM, kNN, and Random Forest achieve the best Micro-F1 score of 0.97.

In terms of classification accuracy, we can see that Random Forest outperforms the baselines as well as the other four classifiers for the first four barriers. Notice that Random forest performs better than decision tree but takes more time. Naive-Bayes achieves a little bit lower classification accuracy than the Decision Tree for the first four barriers. On the political barrier Naive-Bayes achieves the best classification accuracy (0.98) but lower Micro-F1 (0.66).

## 6 CONCLUSIONS AND FUTURE WORK

It is highly important to detect the barriers while information propagates specifically through the news. For journalists, marketers, and social scientists, the phenomenon of knowing which barrier appeared most frequently for what type of events, is significantly helpful to solve business and marketing problems. In this regard, we proposed a simple methodology. Though its results are good enough for three types of events, we would like to enhance features as well as events. We used only Wikipedia-concepts and meta information to detect barriers. In the future, we would like to use DMoz categories provided by Event Registry [10], and transformation of the text of news articles as a feature for barrier detection. Currently geographical and time zone barriers are calculated in a binary way either the same or different. In the future, we would like to introduce the distance between countries and between time zones as labels instead of the currently used binary labeling.

## 7 ACKNOWLEDGMENTS

The research described in this paper was supported by the Slovenian research agency under the project J2-1736 Causalify and co-financed by the Republic of Slovenia and the European Union's Horizon 2020 research and innovation program under the Marie Sk-lodowska-Curie grant agreement No 812997.

**Table 5.** Classifiers' comparison with baselines

| Barrier | Model | CA | Mic-Pre | Mic-Rec | Mic-F1 |
|---|---|---|---|---|---|
| **Economic** | Uniform | 0.50 | 0.50 | 0.49 | 0.49 |
| | Stratified | 0.58 | 0.59 | 0.57 | 0.59 |
| | Most Frequent | 0.70 | 0.70 | 0.70 | 0.70 |
| | SVM | 0.66 | 0.69 | 0.69 | 0.69 |
| | kNN | 0.70 | 0.75 | 0.75 | **0.75** |
| | Decision Tree | 0.69 | 0.73 | 0.73 | 0.73 |
| | Random Forest | 0.74 | 0.74 | 0.74 | 0.74 |
| | Naive Bayes | 0.61 | 0.63 | 0.63 | 0.63 |
| **Cultural** | Uniform | 0.50 | 0.50 | 0.49 | 0.50 |
| | Stratified | 0.83 | 0.83 | 0.83 | 0.83 |
| | Most Frequent | 0.90 | 0.90 | 0.90 | 0.90 |
| | SVM | 0.84 | 0.93 | 0.93 | 0.93 |
| | kNN | 0.55 | 0.95 | 0.95 | **0.95** |
| | Decision Tree | 0.90 | 0.94 | 0.94 | 0.94 |
| | Random Forest | 0.93 | 0.93 | 0.93 | 0.93 |
| | Naive Bayes | 0.83 | 0.51 | 0.51 | 0.51 |
| **Geographical** | Uniform | 0.49 | 0.50 | 0.50 | 0.50 |
| | Stratified | 0.50 | 0.51 | 0.51 | 0.51 |
| | Most Frequent | 0.58 | 0.58 | 0.58 | 0.58 |
| | SVM | 0.81 | 0.76 | 0.76 | 0.76 |
| | kNN | 0.79 | 0.81 | 0.81 | **0.81** |
| | Decision Tree | 0.78 | 0.81 | 0.81 | **0.81** |
| | Random Forest | 0.79 | 0.79 | 0.79 | 0.79 |
| | Naive Bayes | 0.76 | 0.79 | 0.79 | 0.79 |
| **Time Zone** | Uniform | 0.49 | 0.49 | 0.49 | 0.49 |
| | Stratified | 0.59 | 0.58 | 0.58 | 0.58 |
| | Most Frequent | 0.70 | 0.70 | 0.70 | 0.70 |
| | SVM | 0.78 | 0.77 | 0.77 | 0.77 |
| | kNN | 0.70 | 0.78 | 0.78 | 0.78 |
| | Decision Tree | 0.80 | 0.81 | 0.81 | 0.81 |
| | Random Forest | 0.83 | 0.83 | 0.83 | **0.83** |
| | Naive Bayes | 0.72 | 0.64 | 0.64 | 0.64 |
| **Political** | Uniform | 0.51 | 0.52 | 0.50 | 0.50 |
| | Stratified | 0.84 | 0.83 | 0.81 | 0.82 |
| | Most Frequent | 0.90 | 0.90 | 0.90 | 0.90 |
| | SVM | 0.79 | 0.97 | 0.97 | **0.97** |
| | kNN | 0.62 | 0.97 | 0.97 | **0.97** |
| | Decision Tree | 0.79 | 0.91 | 0.91 | 0.91 |
| | Random Forest | 0.97 | 0.97 | 0.97 | **0.97** |
| | Naive Bayes | 0.98 | 0.66 | 0.66 | 0.66 |